\documentclass{article}

\usepackage{cite}
\usepackage{booktabs} 
\usepackage{caption} 
\usepackage{subcaption} 
\usepackage{graphicx}
\usepackage[utf8]{inputenc}
\usepackage{tikz}
\usetikzlibrary{er,positioning,bayesnet}
\usepackage{multicol}
\usepackage{algpseudocode,algorithm,algorithmicx}

\newcommand{\conf}[2]{{ \tiny({#1}, {#2})}}

\usepackage{multirow}

\newcommand{\appendixtext}{Appendix}
\usepackage{indentfirst}
\usepackage{url}

\usepackage{amssymb,amsthm,amsmath}
\usepackage{xcolor,paralist,fancyhdr,etoolbox}

\usepackage{lipsum}

\begin{document}
\title{Prevalence and prevention of large language model use in crowd work} 
\author{Veniamin Veselovsky,$^{1, \dagger}$
Manoel Horta Ribeiro,$^{1, \dagger}$
Philip Cozzolino,$^{2}$\\
Andrew Gordon,$^2$
David Rothschild,$^3$
Robert West$^1$}

\date{}

\maketitle

\begin{center}
    \thanks{$^1$ EPFL; $^2$ Prolific; $^3$ Microsoft Research;\\ $^\dagger$ Equal contribution.}
\end{center}

\let\thefootnote\relax

\begin{abstract}
We show that the use of large language models (LLMs) is prevalent among crowd workers, and that targeted mitigation strategies can significantly reduce, but not eliminate, LLM use.
On a text summarization task where workers were not directed in any way regarding their LLM use, the estimated prevalence of LLM use was around 30\%, but was reduced by about half by asking workers to not use LLMs and by raising the cost of using them, e.g., by disabling copy-pasting.
Secondary analyses give further insight into LLM use and its prevention:
LLM use yields high-quality but homogeneous responses, which may harm research concerned with human (rather than model) behavior and degrade future models trained with crowdsourced data.
At the same time, preventing LLM use may be at odds with obtaining high-quality responses;
e.g., when requesting workers not to use LLMs, summaries contained fewer keywords carrying essential information.
Our estimates will likely change as LLMs increase in popularity or capabilities, and as norms around their usage change. Yet, understanding the co-evolution of LLM-based tools and users is key to maintaining the validity of research done using crowdsourcing, and we provide a critical baseline before widespread adoption ensues.
\end{abstract} 

\bigskip

Crowd work platforms, such as Prolific and Amazon Mechanical Turk, are central in academia and industry, empowering the creation, annotation, and summarization of data~\cite{gray2019ghost}, as well as surveys and experiments~\cite{salganik2019bit}.
At the same time, large language models (LLMs), such as ChatGPT, PaLM, and Claude, promise similar capabilities.
They are remarkable data annotators~\cite{Gilardi2023ChatGPTOC} and can, in some cases, accurately simulate human behavior, enabling \textit{in silico} experiments and surveys that yield human-like results~\cite{argyle2023out}.
Yet, if crowd workers were to start using LLMs, this could threaten the validity of data generated using crowd work platforms. Sometimes, researchers seek to observe unaided human responses (even if LLMs could provide a good proxy), and LLMs still often fail to accurately simulate human behavior~\cite{santurkar2023whose}. 
Further, LLM-generated data may degrade subsequent models trained on it~\cite{shumailov2023model}.
Here, we investigate the extent to which crowd workers use LLMs in a text-production task and whether targeted mitigation strategies can prevent LLM usage.

\section{Study \#1: Prevalence of LLM use}

To estimate LLM usage on Prolific, a research-oriented crowd work platform, we asked $n=168$  workers on 3 July 2023 to summarize scientific abstracts (following \cite{horta2019message}; see \appendixtext{}). 
We chose this task because it is laborious for humans but easily done by LLMs~\cite{luo2023chatgpt} and because we can use pre-LLM summaries from prior work \cite{horta2019message} as `human ground truth'.
We detected whether a summary had been generated using LLMs with a finetuned \texttt{e5-base} classifier~\cite{wang2022text} trained on human pre-LLM summaries~\cite{horta2019message} and summaries generated by GPT-4 and ChatGPT.
The model was then run on each of the 168 new summaries to estimate its probability of being LLM-generated. In this study, we did not direct the participants regarding LLM use in any way, thus capturing a baseline of LLM use for uninstructed participants doing a task for which LLMs have a considerable advantage over human labor.

Inspired by a study on Mechanical Turk from 1--1.5 months before ours~\cite{veselovsky2023artificial}, we used three approaches to aggregate the probabilities, obtaining similar (but slightly lower) estimates:
(1)~\textit{`classify-and-count',} considering as synthetic any summary with an LLM probability above 50\% (prevalence estimate: 33.3\%; 95\% CI $[25.9\%,40.1\%]$);
(2)~\textit{`probabilistic classify-and-count'}, where we calibrated the model \cite{card2018importance} (see \appendixtext{}) and then averaged the LLM probabilities (estimate: 35.2\% $[29.8\%,40.6\%]$);
(3)~\textit{`corrected classify-and-count',} adjusting for the type I and type II error rates estimated on the training data \cite{meyer2017misclassification} (estimate: 35.4\% $[27.8\%,43.0\%]$).

We validated our results by analyzing crowd workers' copy-pasting behavior (see \appendixtext{}{}), finding that 55\% of summaries where workers had copied text were classified as synthetic (i.e., LLM probability above 50\%), vs.\ only 9\% when workers had not copied text.
As no information about copy-pasting was used in the LLM-or-not classifier, this result strengthens our confidence in the classifier.
Interestingly, far fewer crowd workers used copy-pasting on Prolific (53\%) in our Study \#1, compared to a previous study~\cite{veselovsky2023artificial} on Amazon Mechanical Turk (89\%).

\begin{table}[t]
    \small
    
    \begin{subtable}[h]{\linewidth}
\centering
\begin{tabular}{l@{\hskip 1mm}ll@{\hskip 1mm}l@{\hskip 1mm}l}
& & \multicolumn{3}{c}{ \textbf{Hurdle}} \\
\cline{3-5}\addlinespace[.5ex]
  &  & None & Image & Ctrl C+V \\
  \cline{3-5}\addlinespace[.5ex]
\multirow{3}{*}{\rotatebox[origin=c]{90}{\textbf{Request}}} & \multicolumn{1}{|r|}{None} & 27.6\%\conf{21.0\%}{34.6\%} & 21.5\%\conf{16.0\%}{27.4\%} & 24.1\%\conf{18.3\%}{30.4\%} \\
& \multicolumn{1}{|r|}{Indirect}  & 28.5\%\conf{21.7\%}{35.8\%} & 19.8\%\conf{14.2\%}{25.8\%} & 19.3\%\conf{14.6\%}{24.5\%} \\
& \multicolumn{1}{|r|}{Direct}  & 24.0\%\conf{18.6\%}{29.6\%} & 15.9\%\conf{11.9\%}{20.3\%} & 15.8\%\conf{11.8\%}{20.4\%} \\
\cline{3-5}\addlinespace[.5ex]
\end{tabular}
\subcaption{Classifier}
\label{tabsub:p}
\end{subtable}

\begin{subtable}[h]{\linewidth}
\centering
\begin{tabular}{l@{\hskip 1mm}ll@{\hskip 1mm}l@{\hskip 1mm}l}
& & \multicolumn{3}{c}{ \textbf{Hurdle}} \\
\cline{3-5}\addlinespace[.5ex]
  &  & None & Image & Ctrl C+V \\
  \cline{3-5}\addlinespace[.5ex]
\multirow{3}{*}{\rotatebox[origin=c]{90}{\textbf{Request}}} & \multicolumn{1}{|r|}{None} & 15.8\%\conf{8.5\%}{24.4\%} & 10.4\%\conf{3.9\%}{16.9\%} & 4.9\%\conf{1.2\%}{9.8\%} \\

& \multicolumn{1}{|r|}{Indirect}  & 13.2\%\conf{5.9\%}{22.1\%} & 6.6\%\conf{1.3\%}{12.0\%} & 3.6\%\conf{0.0\%}{8.3\%} \\
& \multicolumn{1}{|r|}{Direct}  & 3.0\%\conf{0.0\%}{7.1\%} & 6.6\%\conf{1.3\%}{13.2\%} & 9.1\%\conf{3.9\%}{15.6\%} \\
\cline{3-5}\addlinespace[.5ex] 
\end{tabular}
\subcaption{Self-reported}
\label{tabsub:s}
\end{subtable}

\begin{subtable}[h]{\linewidth}
\centering
\begin{tabular}{l@{\hskip 1mm}ll@{\hskip 1mm}l@{\hskip 1mm}l}
& & \multicolumn{3}{c}{ \textbf{Hurdle}} \\
\cline{3-5}\addlinespace[.5ex]
  &  & None & Image & Ctrl C+V \\
  \cline{3-5}\addlinespace[.5ex]
\multirow{3}{*}{\rotatebox[origin=c]{90}{\textbf{Request}}} & \multicolumn{1}{|r|}{None} & 10.9\%\conf{4.9\%}{18.3\%} & 2.6\%\conf{0.0\%}{6.5\%} & 1.2\%\conf{0.0\%}{3.7\%} \\
& \multicolumn{1}{|r|}{Indirect}  & 4.4\%\conf{0.0\%}{10.3\%} & 5.3\%\conf{1.3\%}{10.7\%} & 2.4\%\conf{0.0\%}{6.0\%} \\
& \multicolumn{1}{|r|}{Direct}  & 7.1\%\conf{3.0\%}{12.1\%} & 4.0\%\conf{0.0\%}{9.2\%} & 0.0\%\conf{0.0\%}{0.0\%} \\
\cline{3-5}\addlinespace[.5ex] 
\end{tabular}
\subcaption{Heuristics}
\label{tabsub:h}
\end{subtable}



    \caption{LLM usage across experimental conditions, estimated using three methods: 
    (a)~probabilistic classify-and-count (`Classifier');
    (b)~self-reported usage (`Self-report');
    (c)~high-precision heuristics (`Heuristics'). 
    All estimates indicate that the interventions significantly reduced LLM usage, albeit not completely.
    }
    \label{tab:interventions_adj}
\end{table}

\section{Study \#2: Prevention of LLM use}

Next, we analyzed whether targeted strategies can curb LLM usage.
Specifically, we studied two different mitigation approaches: (1) explicitly asking crowd workers not to use LLMs (henceforth the `\textit{request}' strategy) and (2) imposing hurdles that deter LLM use (the `\textit{hurdle}' strategy). 
We considered two variations for each: for the \textit{request} strategy, we asked individuals both indirectly and directly not to use LLMs (see \appendixtext{}{}), and for the \textit{hurdle} strategy, we either converted the original abstract text to an image or disabled copy-pasting entirely.
As the two strategies are independent, we
investigated all combinations (alongside a no-restriction condition) in a $3\times3$ factorial design; see Table~\ref{tab:interventions_adj}.

Using the same task and classifier as in Study \#1, Study \#2 was conducted on 23 July 2023, by randomly split $n=720$ users into the nine conditions.
Upon completion, they were then redirected to a follow-up survey where they were asked  
(Q1) how often they used ChatGPT in their daily lives,
(Q2) if they had used ChatGPT for the task, and
(Q3) if they knew of studies tracking ChatGPT usage on crowd work platforms (see \appendixtext{} for exact phrasing).

\subsection{Effectiveness of preventive measures} 
The proposed mitigation strategies significantly reduced LLM usage (Table~\ref{tab:interventions_adj}). 
For example, when workers were directly requested not to use LLMs and shown the text to be summarized as an image (thus preventing copy-pasting), LLM usage (as measured by the probabilistic classify-and-count method) almost halved, dropping from 27.6\% to 15.9\% (Table~\ref{tabsub:p}). 
Similar results were obtained using self-reported usage by crowd workers (Q2) and using high-precision heuristics (Tables~\ref{tabsub:s}--\ref{tabsub:h}; see Materials and Methods).
Comparing high-precision heuristics with self-reports revealed that only 11 of the 31 workers using LLMs according to high-precision heuristics admitted to using LLMs, whereas 31 of the 689 whom the heuristic and classifier both failed to mark as synthetic admitted to LLM usage.

We further disentangle the effect of each specific strategy and variation with a linear model (Fig.~\ref{fig:effects}; cf.\ \appendixtext{}{}).
All variations in both strategies led to decreases in LLM usage across metrics, although this decrease was sometimes not statistically significant when measured individually.
Notably, asking crowd workers indirectly (``Please do your best to summarize the abstract in your own words'') was the least effective strategy across all measures of LLM use and the only non-significant measure when considering the classifier-based outcome (`Indirect'; 2\% decrease, $p=0.38$).
This hints at the complexity of preventing LLM use, which is adversarial in nature. 

\subsection{Correlates of LLM use} We study the relationship between LLM usage and (1) the age of crowd workers and (2) how they answered two of the post-survey questions (awareness of studies measuring LLM usage; LLM usage in general) using a simple linear model and considering both self-reports and the classifier's LLM-probability estimates as outcomes (see \appendixtext{}{}).
We find that younger individuals were significantly more likely to use LLMs ($-0.18\%$ in estimated LLM probability per year; $p=0.014$) and that workers who used LLMs `often' were 18.7\% more likely to use it for the task ($p<0.001$). 
Awareness of studies measuring LLM use did not significantly impact usage (+1.6\%, $p=0.55$).
Results were similar when considering self-reported usage as the outcome variable.

\subsection{Content-level analysis}

Analyzing the text of crowd workers' summaries, we found that summaries labeled as synthetic by the classifier were significantly more \textit{`homogeneous'} than those labeled as human, according to a previously proposed homogeneity metric
 \cite{padmakumar2023does}
 (details in \appendixtext{}{}), with homogeneity score
45.6\% [43.2\%, 48.2\%] for synthetic, vs.\ 
 27.1\% [26.8\%, 27.4\%] for human, summaries.

\begin{figure}[t]
    \centering
    \includegraphics[width=0.6\linewidth]{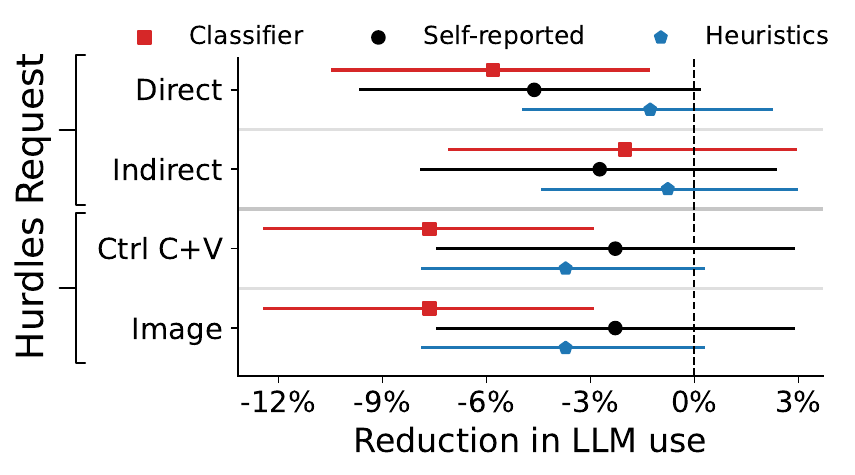}
    \caption{Estimated effect sizes for interventions to prevent LLM usage considering three different measures of LLM use as the outcome variable: (1)~high-precision heuristics, (2)~self-reported usage, (3)~probabilistic classify-and-count.}
    \label{fig:effects}
\end{figure}

In the original study whose human summaries we reused~\cite{horta2019message}, the authors measured the retention of keywords from the original abstract corresponding to essential information, finding it to be highly correlated with human evaluations of quality.
Using this metric as a proxy for quality, we found that summaries labeled as synthetic preserved more keywords 
(40.1\% [36.9\%, 43.2\%]) than summaries labeled as human (31.2\% [29.9\%, 32.6\%]).
We found a similar effect when using self-reports and high-precision heuristics instead of the classifier's labels. 
Finally, we repeated the analysis of Fig.~\ref{fig:effects}, but now considering only summaries labeled as human by the classifier and using homogeneity and keyword retention as outcomes. We found no significant effect of the interventions on either, with one exception: directly requesting workers not to use LLMs decreased keyword retention by 6.2\% ($p=0.009$).
(Further discussion in \appendixtext{}{}).

\section{Discussion}
The results suggest that LLMs pervade current crowd work on text production tasks. Although adopting various strict mitigation approaches reduced LLM use by nearly 50\%, it could not fully prevent it.
Synthetic data may harm the utility of crowd work platforms, as researchers often care about truly human behavior or preferences; e.g., the authors of the paper whose human summaries we borrowed \cite{horta2019message} wanted to know how \emph{people} summarized, rather merely obtain good summaries.
While some preliminary studies suggest that synthetic data may capture certain viewpoints~\cite{argyle2023out}, they still often fail, and research using crowd work may inadvertently capture the behavior and preferences of LLMs, not humans.
Even if LLMs can capture average behavior or preferences, the homogeneity of their responses may result in a loss of the long tail of human behavior and preferences that is vital to researchers~\cite{song2023reward} and, according to recent work, important to training capable LLMs~\cite{shumailov2023model}.

We must be careful not to conflate LLM use with cheating. 
Depending on the study, it could be beneficial if LLMs assist crowd workers.
Further, as LLMs become intertwined with how people write and accomplish everyday tasks, the distinction between `synthetic' and `human' data may blur. 
For example, is text generated with the help of a spellchecker `synthetic'? 
Thus, we expect the thresholds for concern and meaning will shift dramatically over the coming months and years as LLMs become more ubiquitous in everyday productivity tasks. 
In that context, a fruitful future direction is to explore the landscape of \emph{how} crowd workers use LLMs. 
There are many ways of integrating these models into crowd workers' workflows, and different approaches may have different effects on downstream research output. 

We found that stricter mitigation approaches can significantly reduce LLM use. 
These measures may, however, backfire when detection is critical. 
Stricter measures may limit the number of participants using LLMs but also make them more reluctant to admit ex-post that they used them, or make them harder to detect, as the prevention measure eliminated a key indicator of LLM use. 
For example, eliminating copy-pasting makes it harder to use LLMs, limiting use, but then researchers also cannot use copy-pasting as a feature to detect who used LLMs.
Further, mitigation approaches can reduce the overall response quality: as we found empirically, workers explicitly told not to use LLMs produced lower-quality summaries.

LLM-based tools and LLM users are co-evolving in ways to ensure the low temporal validity of our specific findings and estimates. 
In the last few months, tools have evolved to interpret images and to 
call LLMs without the need to copy-paste (e.g., by simply selecting text). 
This does not diminish the value of our work; it makes it even more valuable: 
it is critical to establish baselines and ongoing measurements as this co-evolution progresses, and and our work establishes such  baselines. 
Further, we are confident that our high-level interpretations and guidance will translate across this evolution, and we hope this helps establish a regularly updated new program of study to serve crowd work platforms as well as researchers.

\section{Materials and Methods}
\subsection{Data}
We modified a prior Mechanical Turk task  \cite{horta2019message} where crowd workers were asked to summarize medical paper abstracts. We re-ran the study twice on Prolific. In Study \#1, we estimated prevalence by collecting 168 user summaries (paid £9/hour). In Study \#2, we re-ran the study on 720 users, now using several mitigation techniques (paying £10/hour). (Full description of data and original study in \appendixtext{}{}.)

\subsection{Model training}
We finetuned a \texttt{e5-base-v2} language model  \cite{wang2022text} for our classification task and conducted a hyperparameter sweep. 
The model was trained on the summaries from the original study~\cite{horta2019message} (written before the adoption of LLMs) and summaries synthetically generated using OpenAI's API. 

\subsection{Heuristic-based estimates}
We defined two high-precision heuristics for measuring LLM use: feasible time for completion and pasting in artifacts from the ChatGPT Web interface. (Details in \appendixtext{}{}).

\subsection{Effect of each intervention} We assessed the effectiveness of each of the interventions with a linear probability model. We do not consider interactions between the treatment conditions, as a two-way ANOVA indicated that the interactions between the two strategies are not statistically significant. 

\bibliographystyle{plain}

\bibliography{temp}

\appendix

\section{Data}\label{data}

\vspace{2mm}
\noindent
\textbf{Overall approach.}
For both studies, we modify a prior MTurk task originally devised by Horta Ribeiro et al.~\cite{horta2019message}, whose goal was to study the so-called ``telephone effect,'' whereby information is gradually lost or distorted as a message is passed from human to human in an information cascade.
Specifically, in that study, crowd workers were asked to summarize a research abstract.
We chose this task for our study for two reasons. 
First, it is laborious for humans while being easily done with the aid of commercially available LLMs~\cite{luo2023chatgpt}. Second, it is a good example of a task where truly human text is fundamentally required: the very point of \cite{horta2019message} was to study how information is lost when \textit{humans} summarize text, which would not have been possible with synthetically generated, rather than human-generated, data.
In the original study, crowd workers produced eight increasingly short summaries of each original abstract, forming entire information cascades. For our purpose, however, we reduced the task to a single summarization step, where an abstract was condensed into a concise summary of ideally about 100 words.

\vspace{2mm}
\noindent
\textbf{Synthetic data.} We generate synthetic summaries of the abstracts using ChatGPT with two prompting approaches. A direct copying of the original instructions and a simpler request (both shown below). We then feed these prompts into gpt-3.5-turbo and GPT-4 to generate 1440 synthetic summaries across the two prompts. We used temperature values of 0.5, 0.7, 0.9, and 1.0 for these generations.

\vspace{2mm}
\noindent
\textbf{Original summaries data.}
In Horta Ribeiro et al.~\cite{horta2019message}, crowd workers were given medical research paper abstracts published in the \textit{New England Journal of Medicine} (NEJM) and asked to summarize the original abstract (about 2000 characters) into shorter versions~(1024, 512, 256, 128, and 56 characters). 
Further, in another experimental condition, participants were asked to do this summarization iteratively, i.e., first, someone would summarize the original abstract into a 1,024-character summary, then, someone else would summarize the 1,024-character summary into 512 characters, and so on.
The original abstracts were about four research topics of public interest (vaccination, breast cancer, cardiovascular disease, and nutrition), and four papers were selected per topic, for a total of 16 abstracts.
Given the increasingly short nature of the summaries, here we only consider the first three summarization levels, i.e., around 1,024, 512, and 256 characters, corresponding to 724 summaries.

\vspace{2mm}
\noindent
\textbf{Data availability.}
We make all data from the study publically available.\footnote{\url{https://github.com/epfl-dlab/LLM-crowdwork/data/}}

\vspace{2mm}
\noindent
\textbf{Prompts.} We consider the two following prompts to generate synthetic texts:

\noindent \dotfill

\noindent
\textbf{Prompt \#1}
\newline
\noindent
Summarize the following abstract: \textbf{\{text\}}

\noindent \dotfill

\noindent
\textbf{Prompt \#2}
\newline
\noindent
You will be given a short text (around 400 words) with medicine-related information.
\newline
\noindent
Your task is to:
\newline
\noindent
Read the text carefully.
\newline
\noindent
Write a summary of the text. Your summary should:
\newline
\noindent
Convey the most important information in the text, as if you are trying to inform another person about what you just read.
\newline
\noindent
Contain at least 100 words.
\newline
\noindent
We expect high-quality summaries and will manually inspect some of them.

Text: \textbf{\{text\}}

\noindent \dotfill

\section{Experimental details}

In study \#1, participants were asked to summarize one out of 16 medical abstracts. In study \#2, we asked participants to execute the same task, but participants were randomly sorted into nine different experimental conditions, which resulted from a combination of three ways of requesting the summary (`None', `Indirect', `Direct'), and three ways of making LLM use harder (`None', `Image', `No Ctrl C+V'). 
For the hurdles to stop LLM-use, we make available the images containing each summary as well as the HTML code preventing copy-pasting.\footnote{\url{https://github.com/epfl-dlab/LLM-crowdwork/experiments/}}

\vspace{2mm}
\noindent
\textbf{Instructions.}
We reproduce the instructions for studies \#1 and \#2 below. For study \#2, we show the texts associated with the three request conditions. 

\noindent \dotfill

\noindent
\textbf{Instructions for study \#1}
\newline
\noindent
You will be given a short text (around 400 words) with medicine-related information.

\noindent
Your task is to:

\begin{itemize}
    \item Read the text carefully.
    \item Write a summary of the text. Your summary should:
\begin{itemize}
    \item Convey the most important information in the text, as if you are trying to inform another person about what you just read.
    \item Contain at least 100 words.
\end{itemize}
\end{itemize}

\noindent
Contain at least 100 words.
We expect high-quality summaries and will manually inspect some of them.

\noindent \dotfill

\noindent
\textbf{Instructions for study \#2}
\newline
You will be given a short text (around 400 words) with medicine-related information.

\noindent
Your task is to:

\begin{itemize}
    \item Read the text carefully.
    \item Write a summary of the text. Your summary should:
\begin{itemize}
    \item Convey the most important information in the text, as if you are trying to inform another person about what you just read.
    \item Contain at least 100 words.
\end{itemize}
\end{itemize}

\noindent
Contain at least 100 words.
\textbf{\{Request-related text.\}}

\noindent \dotfill

\noindent
\textbf{Request condition}: `None.' 
\\\textbf{Text:} We expect high-quality summaries and will manually inspect some of them.

\vspace{2.5mm}

\noindent
\textbf{Request condition}: `Indirect.' \\ 
\textbf{Text:} For this task, we are looking for the \textbf{best human generated responses possible}. Even though the material may be unfamiliar to you, please do your best to summarize the abstract in your own words.

\vspace{2.5mm}

\noindent
\textbf{Request condition}: `Direct.' \\ 
\textbf{Text:} For this task, \textbf{it is important that you do not use ChatGPT (or other AI models) to generate the abstract summaries}. Even though the material may be unfamiliar to you, do not rely on AI, as we will be monitoring to see if you use those tools.

\noindent \dotfill

\noindent 
\textbf{Follow-up questions.} For study 2, we also ask a series of follow-up questions. 

\vspace{2.5mm}

\noindent
Thanks for completing this task. We would like you to answer a couple of questions about AI-usage.
If you did use AI tools on the previous task you will still be compensated after answering these questions, even if we said we would be monitoring AI-use.
\textbf{Your Prolific participant account will not be affected negatively based on how you answer these questions!}
            
\begin{enumerate}
    \item Did you use AI tools like ChatGPT to accomplish the previous task? (Yes/No)
    \item Were you aware of previous research investigating AI use among crowd workers? (Yes/No)
    \item How often do you use ChatGPT and similar AI tools in your daily life? (Rarely/Occasionally/Often)
\end{enumerate}

\noindent 
\section{Model training}\label{training}
\noindent 
We release all code used to train the model.\footnote{\url{https://github.com/epfl-dlab/LLM-crowdwork/model/}} 

\vspace{2mm}
\noindent
\textbf{Training set} We train the model on real texts from the original 2019 study, and use texts generated by ChatGPT as synthetic texts (see Section \ref{data}). We then break up the train and test dataset into a train set of 1,733 (653 real, 1080 synthetic) and a test of 431 points (71 real, 360 synthetic). Out of the train points, we set aside 100 random points for validation. 

\vspace{2mm}
\noindent
\textbf{Hyperparameters} A hyperparameter sweep over learning rate, batch size, and gradient accumulation steps was conducted using Optuna~\cite{akiba2019optuna}. After we set the learning rate to 3e-5, trained for 3 epochs, used a weight decay of 0.01, a warmup ratio of 0.1, and a batch size of 32 (precisely, a batch size of 8 with 4 gradient accumulation steps).

\vspace{2mm}
\noindent
\textbf{Results} We trained the model five times using different seeds. All models achieved comparable precision recalls on the validation sets, with macro-f1 being above 97\% across all runs. Precision was always above 99\%, and recall was above 96.5\%. We chose the model with the highest macro-f1 for reporting the results.

\vspace{2mm}
\noindent
\textbf{Model calibration}
Temperature scaling is a post-processing calibration method to improve the confidence estimates of pre-trained neural networks. More specifically, it adjusts the network's confidence in such a manner that the confidence better aligns with the empirical accuracy.  
Temperature scaling works by introducing a scalar parameter T (temperature) that scales the logits (outputs of the last layer pre-softmax in the neural network model) before applying softmax function for classification. Given output logits $z = (z_{1}, ..., z_{K})$ of a neural network model, the probability $p_{i}$ of class $i$ with temperature scaling is computed as follows in a softmax layer:


\begin{align} 
    p_{i}(T) = \frac{e^{z_{i}/T}}{\sum_{j=1}^{K} e^{z_{j}/T}}  \quad \text{for} \quad i = 1, ..., K 
\end{align}

We determine T by minimizing the Expected Calibration Error (ECE). ECE is a metric used to measure the miscalibration of a classification model. It compares the model's predicted confidence for a class to the accuracy of the model in predicting that class. Specifically, it computes the expected difference between the predicted confidence and the actual accuracy.
After calibrating the model, the ECE drops from 0.897 to 0.782.

\section{Prevalence estimation}
\subsection{Aggregating the predictions} Combining model predictions risks over- or under-counting by adding false positives or negatives. Prevalence adjustment techniques aim to account for these biases in the models, which we utilize throughout this paper. 

\vspace{2mm}
\noindent
\textbf{Classify and count (CC)} A synthetic summary is determined as any summary with a probability of being synthetic greater than 0.5. We then calculate prevalence as the fraction of summaries that are predicted summaries. 

\vspace{2mm}
\noindent
\textbf{Calibrated probabilistic classify and count (PCC$^\text{cal}$)} As per~\cite{card2018importance}, we adjust the prevalence metric by summing probabilities on a calibrated version of our classifier. Calibration is done by applying temperature scaling on the output logits to minimize the expected calibration error (see Section \ref{training} for details on calibration).

\vspace{2mm}
\noindent
\textbf{Adjusted Classify and count (Adj)} 
Previous work~\cite{hausman1998misclassification,meyer2017misclassification} has studied how the coefficients of linear probability models can be corrected for misclassification bias on the outcome variable.
Namely, in a generic linear  model:
\begin{align*}
    y =  \beta_0 + \beta_1 x_1 + \ldots x_k \beta_k + \epsilon
\end{align*}
if the outcome $y$ suffers from misclassification with a false positive rate $\alpha_0$, and a false negative rate $\alpha_1$,
the coefficients estimated will be biased. However, this bias can be easily corrected, as the it can be shown that:

\begin{align*}
    \mathbf{E}(\hat{\beta}_0) &= \alpha_0 +(1 - \alpha_0 - \alpha_1)\beta_0 \\
    \mathbf{E}(\hat{\beta}_{1\ldots k}) &= (1 - \alpha_0 - \alpha_1)\beta_0
\end{align*}

Using this intuition, we use the training set to estimate the false positive and false negative rates and correct the classify and count estimators using the formulas above.


\subsection{Heuristics: lower bound for use} We construct two high-precision, low-recall heuristic to act as strict-lower bounds for LLM-use.  

\vspace{2mm}
\noindent
\textbf{Chat interface artifacts} ChatGPT web interface adds four new lines to the bottom of the text, when it’s copied a specific way –– by extending the cursor into the next textbox. We track if pasted in text includes these new lines. 

\vspace{2mm}
\noindent
\textbf{Feasible time} Using estimates of unlikely reading and typing speeds we will estimate an infeasible timeframe to complete the task. The selected threshold will be a reading speed at 400 words per min (wpm) and typing speed at 80 wpm. If someone completes the assignment in less than this time conditioning on the length of the task and abstract, and length of written response.

\section{Content-level analysis}
\subsection{Homogeneity}
We adopt the approach from~\cite{padmakumar2023does} to measure the diversity of the summaries predicted real and synthetic. In particular, for study \#2 we break up the summaries of each abstract into those predicted real and those predicted synthetic by the classifier, heuristics, and self-reports. We then calculate the rouge-L score between all the summaries in a given group and aggregate for that group across all summaries. Precisely, we define the homogeneity  score for a specific abstract as follows: 
\begin{align}
    \text{Homogeneity}(i,c) = \frac{1}{|S_i|}\sum_{j \in S_i^c}\sum_{k \in {S_i^c}: k \neq j} \text{lcs}(s_{ij}, s_{ik})
\end{align}

Where $i$ is the index of an abstract, and $S_i$ is the set of summaries for that abstract. Then, $c$ is the set of abstracts predicted $c$ (e.g., synthetic), making up a subset $A_i^c$. We then take the longest common subsequence between two distinct summaries $s_{ij}$ and $s_{ik}$. Then, we take the average over the abstracts to get the homogeneity for a prediction class. 

\subsection{Key token survival}
To measure the quality of the summaries, we use an approach proposed by the paper from where we obtained the original summaries~\cite{horta2019message}, namely, for each of the 16 research abstracts, we highlight key tokens that are associated with the 1) participants of the study; 2) intervention in the study; and 3) outcomes of the study. 
Note that this metric favors extractive over abstractive summarization---changing the text incurs the risk of changing key tokens.
Yet, on average, summaries predicted real have larger longest common substrings than the original summaries, suggesting quality might be underestimated for LLM-use with this metric. Given that we find an effect regardless of this, we consider using this metric appropriate.

\noindent
We present one of the 16 summaries with the key tokens underscored below. 

\noindent \dotfill

\noindent
\textbf{A Population-Based Study of Measles, Mumps, and Rubella Vaccination and Autism }

\noindent
It has been suggested that \underline{vaccination} against \underline{measles, mumps, and rubella} (\underline{MMR}) is a cause of \underline{autism}. We conducted a retrospective cohort study of all 
\underline{child}ren born in \underline{Denmark} from January 1991 through December 1998. The cohort 
was selected on the basis of data from the Danish Civil Registration System, 
which assigns a unique identification number to every live-born infant and new
resident in \underline{Denmark}. \underline{MMR}-\underline{vaccination} status was obtained from the Danish 
National Board of Health. Information on the \underline{child}ren's \underline{autism} status was 
obtained from the Danish Psychiatric Central Register, which contains 
information on all diagnoses received by patients in psychiatric hospitals and
outpatient clinics in \underline{Denmark}. We obtained information on potential confounders
from the Danish Medical Birth Registry, the National Hospital Registry, and 
Statistics Denmark. Of the \underline{537,303} \underline{child}ren in the cohort (representing 
2,129,864 person-years), 440,655 (82.0 percent) had received the \underline{MMR} vaccine. We identified \underline{316} \underline{child}ren with a diagnosis of autistic disorder and \underline{422} with
a diagnosis of other \underline{autistic-spectrum} disorders. After adjustment for 
potential confounders, the relative risk of autistic disorder in the group of 
vaccinated \underline{child}ren, as compared with the \underline{unvaccinated} group, was \underline{0.92} (95 
percent confidence interval, 0.68 to 1.24), and the relative risk of another
\underline{autistic-spectrum} disorder was \underline{0.83} (95 percent confidence interval, 0.65 to 
1.07). There was no association between the age at the time of \underline{vaccination}, 
the time since \underline{vaccination}, or the date of \underline{vaccination} and the development of 
autistic disorder. This study provides strong evidence against the hypothesis 
that \underline{MMR} \underline{vaccination} causes \underline{autism}.

\noindent \dotfill

\section{Robustness checks}

\noindent
\textbf{GPTZero.} We validate the predictions of our model with those of GPTZero, a paid synthetic text detector.\footnote{\url{https://gptzero.me/}} Using the API access (API version 2023-05-23), we feed in all the summaries from the mitigation study. 
We use CC for estimates and denote the predictions of the two models as $p(+,+)$ when they both predict the text to be synthetic and $p(-,-)$ when they predict both real. 
Then $p(-,+)$ when our model predicts a text as real and GPTZero predicts a text as synthetic, $p(+,-)$ captures the opposite. We find that $p(+,+)=0.62$, $p(-,-)=0.24$, $p(+,-)=0.095$, and $p(-,+)=0.05$. In summary, both models tend to align on most real texts, but they misalign more on the texts they predict as being synthetic. Using CC estimates, GPTZero predicts 28.5\% of the summaries in study \#1 as synthetic (as opposed to 33.3\% using our model).

\vspace{1mm}
\noindent
\textbf{Ensemble.}
Additionally, we combine the two models into one ensemble model by taking an equal weighting of the outputted probabilities. This ensemble model gives an estimate of 28.4\% of crowd workers using LLMs in our study \#1. However, GPTZero was uncalibrated and overly confident in real texts, rendering it difficult to combine the models meaningfully.

\vspace{1mm}
\noindent
\textbf{Copy-pasting validation.} In study \#1, we further validate our model by analyzing copy-pasting behavior on the task. This was done by logging keystrokes when participants had their browsers focused on the HTML page where they were summarizing the research abstracts. We find that summaries where workers copied were predicted synthetic 55\% of the time, whereas when they did not copy, they were predicted synthetic only 9\%. 

\vspace{1mm}
\noindent
Copied text, however, may simply be a form of extractive summarization. To explore this, we measure what fraction of the summary, when copied, came from the original abstract. 71\% of the real texts vs. 91\% of synthetic summaries shared less than 20\% of the original summary. This means that real texts included longer form extractive summarizations 20\% more of the time. Moreover, no synthetic summary shared more than 40\% similarity with the original abstract, whereas 5 of the real summaries shared more than 50\%.

\section{Statistical methods}

\subsection{Effect of each intervention} 
We disentangle the effectiveness of each intervention with a linear probability model:
\begin{align*}
        y_i = \beta_0 &+
    \beta_1 \cdot \text{Subtle}_i + \beta_2 \cdot \text{Strict}_i  + \beta_3 \cdot \text{Image}_i + \beta_4 \cdot \text{No-copying}_i + \epsilon_i, 
\end{align*}
where `$\text{Subtle}_i$', `$\text{Image}_i$', etc., are indicator variables corresponding to participant $i$'s experimental condition and $y_i$ is the proxy metric for LLM usage. We do not consider interactions between the treatment conditions as a 2-ways ANOVA indicates that the interactions between the two strategies are not statistically significant.
\newline
\subsection{Correlates with LLM-use} 
To study the relationship between LLM use in the task and 1) age (an integer), 2) overall LLM use (never/sometimes/often), and 3) awareness of studies measuring LLM use (yes/no), we use a simple linear probability model:
\begin{align*}
        y_i = \beta_0 +
    \beta_1 \cdot \text{Age}_i &+ 
    \beta_2 \cdot \mathbf{1}\{\text{Awareness}_i=\text{yes}\} \\
    &+ \beta_3 \cdot \mathbf{1}\{\text{LLMUse}_i=\text{sometimes}\} \\
    &+ \beta_4 \cdot \mathbf{1}\{\text{LLMUse}_i=\text{often}\}\\
    &+ \epsilon_i, 
\end{align*}
where `$\text{Age}_i$' the age of participant $i$;
`$\mathbf{1}\{\text{Awareness}_i=\text{yes}\}$' is an indicator variable capturing awareness of studies measuring LLM use (obtained from the questionnaire);
and
`$\mathbf{1}\{\text{LLMUse}_i=\text{*}\}$' are indicator variables capturing overall LLM use (also obtained from the questionnaire). Last, $y_i$ is the proxy metric for LLM usage.

\end{document}